\documentclass{acm_proc_article-sp}

\usepackage{fadi}
\usepackage{fadiEn}
\usepackage{url}

\makeatletter
\let\@copyrightspace\relax
\makeatother

\begin{document}

\title{{\ttlit EdHibou}: a Customizable Interface\\ 
for Decision Support in a Semantic Portal}
%
%
%
%
%

\numberofauthors{4} 
%
\author{
%
%
\alignauthor
Fadi Badra\\
       \affaddr{LORIA (UMR 7503 CNRS--INPL--INRIA-Nancy~2--UHP)}\\ 
       \affaddr{Vand{\oe}uvre-l{\`es}-Nancy, France}\\
       \email{badra@loria.fr}
\alignauthor
Mathieu d'Aquin\\
       \affaddr{Knowledge Media Institute}\\
       \affaddr{The Open University}\\
       \affaddr{United Kingdom}\\
       \email{m.daquin@open.ac.uk}
\alignauthor 
Jean Lieber\\
       \affaddr{LORIA (UMR 7503 CNRS--INPL--INRIA-Nancy~2--UHP)}\\ 
       \affaddr{Vand{\oe}uvre-l{\`es}-Nancy, France}\\
       \email{lieber@loria.fr}
\and  
\alignauthor Thomas Meilender\\
       \affaddr{LORIA (UMR 7503 CNRS--INPL--INRIA-Nancy~2--UHP)}\\ 
       \affaddr{Vand{\oe}uvre-l{\`es}-Nancy, France}\\
       \email{meilendt@loria.fr}
}
\date{25 July 2008}

\maketitle
\begin{abstract}
The Semantic Web is becoming more and more a reality, 
as the required technologies have reached an appropriate level of
maturity. 
However, at this stage, it is important to provide tools facilitating 
the use and deployment of these technologies by end-users. 
In this paper, we describe EdHibou, an automatically generated, 
ontology-based graphical user interface that integrates in a semantic
portal. 
The particularity of EdHibou is that it makes use of OWL reasoning
capabilities 
to provide intelligent features, such as decision support, 
upon the underlying ontology. 
We present an application of EdHibou to medical decision support 
based on a formalization of clinical guidelines in OWL 
and show how it can be customized thanks to an ontology of graphical
components. 
\end{abstract}




\section{Introduction}



The $\kasimir$ project is a multidisciplinary project which aims at
providing oncology practitioners of the Lorraine region of France with
decision support and knowledge management tools.
The $\kasimir$ system is a clinical decision support system which
relies on the formalization of a set of clinical guidelines issued by
the regional health network.   
It uses decision knowledge contained in an OWL ontology
to provide decision support to clinicians.
%
In such an ontology $\Ontologie$, a class $\Patient$ denotes the class
of all patients,  
a class $\Traitement$ denotes the class of all treatments
and a property $\reco$ links a class of patients 
to a class of recommended treatments.
Then to a class $\fm{P}$ of patients is associated a treatment $\fm{T}$ by
 an axiom 
\begin{equation}\label{decisionknowledge}
\fm{P} \subsumepar \exists\reco.\fm{T}
\text{~~ where ~~}
\begin{cases}
\fm{P} \subsumepar \Patient\\ 
\fm{T} \subsumepar \Traitement
\end{cases}\\
\end{equation}

A medical situation is represented by an instance 
$\instance$ of the class $\fm{Patient}$ in the 
ontology $\Ontologie$.
The system then exploits axioms of the form 
(\ref{decisionknowledge}) to associate a set of recommended 
treatments to the patient represented by $\instance$.
Deciding which treatments to recommend to the patient
represented by $\instance$  amounts to finding the most specific atomic concepts $\fm{T}$
 in $\Ontologie$ such that 
$\DecouleOntologie~(\exists\reco.\fm{T})(\instance)$ holds.

The original motivation when developing $\edhibou$ was 
to provide a user interface for the $\kasimir$ system that lets 
the user describe a medical situation for 
which a decision has to be taken.
Such a graphical user interface should let the user
complete the description of an OWL instance $\instance$ and 
trigger some reasoning tasks on the underlying 
 OWL representation of clinical guidelines
to propose a set of recommended treatments.
We built EdHibou as a generic framework, allowing application
developers to generate customizable interfaces to ontologies and
ontology reasoning. The Kasimir system takes advantage of this
framework as an application of EdHibou. 
The key idea in $\edhibou$ is to allow the end-user to 
edit an OWL instance without having to manipulate the OWL syntax, 
by simply filling in values in a form.
When developing this application, the main requirements 
were to make it generic --- so that it can be easily reused 
in other applications, and easy to deploy.
It also had to be customizable.

\section{System Architecture}


Our goal in developing $\edhibou$
was to build a lightweight knowledge edition tool
with 
(1)~a very flexible knowledge model, and 
(2)~highly configurable knowledge acquisition forms.
Apart from the dynamic user interface update mechanism,
anything had to be configurable, including the choice of 
the components to display and how they are displayed.
Requirement (1) has been met by externalizing 
the knowledge model to a distant knowledge server.
The role of this knowledge server is to manage 
a knowledge base and perform all reasoning 
tasks over OWL ontologies.
Requirement (2) was fulfilled by 
pushing application configuration 
into an ontology. 
The generation of the user interface 
is then handled by a simple wrapper 
that takes as input an 
automatically generated XML representation 
of the content of an ontology together
with a set of graphical component implementations.

\begin{figure}
\centering
\epsfig{file=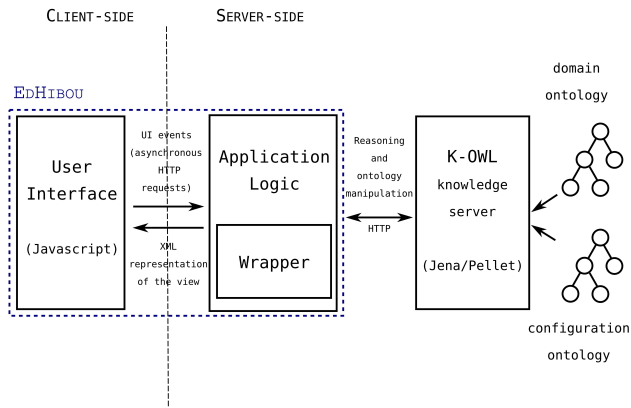,width=8.45cm}
\label{fig:archi}
\caption{$\edhibou$'s software architecture.}
\end{figure}

$\edhibou$ implements a Model-View-Controller architecture pattern
(see figure \ref{fig:archi}) 
and was developed using the Google Web Toolkit 
Java AJAX programming framework.
K-OWL, the knowledge server, 
is a standalone component that plays the role of the model. 
Though it manages knowledge, and not persistent data, 
K-OWL has been designed in quite the same spirit as 
standard database management systems. 
It stores a set of Java models of OWL ontologies 
that are created with the Jena Java API coupled to 
the OWL DL reasoner Pellet.
These ontologies are queried upon over HTTP
using the SPARQL-DL query language~\cite{sparqldl}.
Though K-OWL could be used remotely, it has been 
recently integrated to $\edhibou$'s application logic 
for better performances.

\section{An Ontology-Driven Graphical User Interface Generation}
\label{uigeneration}

In $\edhibou$, all configuration information, 
that is information used to render the 
application knowledge model onto the user interface, 
is placed in a separate ontology.
This ontology contains an exhaustive description
of all graphical components used to build the user interface 
---including their Java implementation classes,
as well as some decision knowledge used to 
determine which graphical component to associate to 
each property of the domain ontology.
The application thus manages in a knowledge base 
an implementation-independant model of the user interface, 
and generating the user interface amounts 
to wrapping this model onto the application view.
Such a  wrapper has been implemented.
Its role is to produce a suitable XML representation 
of the graphical components to display
and transmit it to the application view for rebuild.
Explicitly representing in OWL the configuration knowledge 
has many advantages. One of these advantages, 
compared to hard-coding decision procedures in Java classes,
is to allow easy customization 
by a simple ontology extension mechanism.
A default ontology $\interfaceOntoDefault$ is 
provided that contains some default graphical components 
as well as basic decision knowledge.
$\edhibou$ can then be customized for 
particular domain ontology $\domainOnto$
by replacing this default ontology by a customized
ontology $\interfaceOnto$ that extends $\interfaceOntoDefault$.
This new ontology may add new components,
specify which components to use for a particular property
or even add new decision knowledge to change the 
 behavior of the application.
This customization can 
be done by a knowledge engineer by the means 
of some ontology editor (or ultimately, by $\edhibou$,
 as it is itself a knowledge acquisition tool).


The set of graphical components to display is 
determined at runtime according to the description 
of the currently edited individual.
$\edhibou$'s default behavior is to select 
 the set of graphical components to 
be displayed by testing 
the description of the currently edited instance
against the definition domains of the different 
properties present in the ontology.
To determine which components to display,
$\edhibou$ keeps only 
the properties $\fm{p}$ of the ontology 
for which $\instance$ is an 
instance of the domain of $\fm{p}$ (according to the reasoner).

\section{Related Work}

A number of systems have been developed 
with the aim of generating web interfaces 
on the basis of ontologies and RDF data. 
These systems generally consider the broad task of
 creating \emph{semantic portals}, 
that are websites relying on semantic data. 
For example, \textsc{OntoViews}~\cite{ontoviews} 
is a tool to create such a semantic portal, presenting 
information contained in RDFS ontologies and providing 
navigation and search mechanisms within these ontologies. 
Another example is \textsc{ODESeW}~\cite{odesew}, 
which generates complete \emph{knowledge portal} 
dedicated to the publication and management of information 
in an organization. 
%
Compared to \textsc{EdHibou}, these tools are generally focused 
on the use of ontologies for the presentation of data in a website. 
\textsc{ODESeW} also includes a feature that generates 
forms to create and edit instances as a way to populate the portal. 
However, this functionality does not make use of 
the reasoning capability associated with OWL to guide 
the instance editing process or to infer new information 
from the elements entered by the user. For this reason, 
it could not be use as a way to provide advanced features 
exploiting the knowledge contained in the ontologies, 
like it is done in the \textsc{Kasimir} project 
with clinical decision support, thanks to \textsc{EdHibou}.

\section{Conclusion}
$\edhibou$ is a programmatic framework that
enables to edit an OWL instance by the means of some user-friendly
forms. 
It implements an ontology-driven
graphical user interface generation approach 
and enables to exploit the standard
reasoning on the underlying ontologies 
to provide intelligent behavior. 
An application of $\edhibou$ is presented in which it is integrated in
a semantic portal as a user interface for a decision support system in
oncology. A first demo is currently available online at the URI
\url{http://labotalc.loria.fr/Kasimir}. 

%
\bibliographystyle{abbrv}
\bibliography{edhibou,related-work}  
%
%
\balancecolumns
\end{document}